\documentclass{article}

\usepackage[final,nonatbib]{neurips_2020}

\usepackage[utf8]{inputenc} % allow utf-8 input
\usepackage[T1]{fontenc}    % use 8-bit T1 fonts
\usepackage{hyperref}       % hyperlinks
\usepackage{url}            % simple URL typesetting
\usepackage{booktabs}       % professional-quality tables
\usepackage{amsfonts}       % blackboard math symbols
\usepackage{nicefrac}       % compact symbols for 1/2, etc.
\usepackage{microtype}      % microtypography
\usepackage{amsmath}
\usepackage{caption}
\usepackage{epsfig}
\usepackage{relsize}
\usepackage{wrapfig}
\usepackage{lipsum}

\usepackage[numbers]{natbib}
\title{Multi-layered tensor networks for image classification}
\def \Rm{{\mathbb{R}}}

\def \xbf{{\mathbf x}}

\def \0bf{{\mathbf 0}}

\author{%
  Raghavendra Selvan \\
  Department of Computer Science\\
  Department of Neuroscience\\
  University of Copenhagen\\
  Denmark\\
  \texttt{raghav@di.ku.dk}
   \AND
  Silas Ørting \\
  Department of Computer Science\\
  University of Copenhagen\\
  Denmark\\
  \texttt{silas@di.ku.dk}
  \And
  Erik B Dam \\
  Department of Computer Science\\
  University of Copenhagen\\
  Denmark\\
  \texttt{erikdam@di.ku.dk}
}

\begin{document}

\maketitle
%\vspace{-0.65cm}
\begin{abstract}
  The recently introduced locally orderless tensor network (LoTeNet) for supervised image classification uses matrix product state (MPS) operations on grids of transformed image patches. The resulting patch representations are combined back together into the image space and aggregated hierarchically using multiple MPS blocks per layer to obtain the final decision rules. In this work, we propose a non-patch based modification to LoTeNet that performs one MPS operation per layer, instead of several patch-level operations. The spatial information in the input images to MPS blocks at each layer is {\em squeezed} into the feature dimension, similar to LoTeNet, to maximise retained spatial correlation between pixels when images are flattened into 1D vectors. The proposed multi-layered tensor network (MLTN) is capable of learning linear decision boundaries in high dimensional spaces in a multi-layered setting, which results in a  reduction in the computation cost compared to LoTeNet without any degradation in performance.\footnote{Source code is available at \url{https://github.com/raghavian/mltn}}
  %\vspace{-0.25cm}
\end{abstract}

\section{Introduction}
Tensor networks are factorisations of higher order tensors into lower order tensors~\cite{perez2006matrix,oseledets2011tensor}. Of late, such factorised tensor representations have seen an increased interest in supervised learning since the early work that presented connections to machine learning in~\cite{novikov2015tensorizing,stoudenmire2016supervised,novikov2016exponential}. One reason for this interest could be attributed to the possibility of using tensor networks within the end-to-end learning settings enabled by automatic differentiation, which has driven much of current deep learning ~\cite{schmidhuber2015deep,lecun2015deep}.

In this work, we focus on supervised image classification using the tensor trains\footnote{Matrix product states (MPS) and tensor trains are used interchangeably in literature. We adhere to MPS.} or the matrix product states (MPS) tensor networks which factorise any order-N tensor into a chain (network) of lower order tensors ~\cite{perez2006matrix,efthymiou2019tensornetwork,raghav2020tensor}. In~\cite{stoudenmire2016supervised,efthymiou2019tensornetwork}, 2-D images are flattened into 1-D vectors, lifted to exponentially high dimensions and a linear decision boundary is approximated using MPS. One approach to optimise weights of the MPS approximation is using sweeping algorithms similar to the density matrix renormalization group (DMRG) algorithm~\cite{mcculloch2007density} as performed in~\cite{stoudenmire2016supervised}. Another approach is to optimise the MPS weights using gradient based optimisation using automatic differentiation implemented using the backpropagation algorithm, as performed in~\cite{efthymiou2019tensornetwork,raghav2020tensor}.
% Further, image classification has been addressed quite extensively to benefit from tensor networks ~\cite{raghav2020tensor,liu2019machine,reyes2020multi,selvan2020locally,cheng2020supervised}.

The flattening of 2-D images into 1-D vectors results in loss of spatial correlation between pixels. This was addressed for medical image classification in the locally orderless tensor network (LoTeNet) introduced for 2-D images in~\cite{raghav2020tensor} and extended to volumetric 3-D data in~\cite{selvan2020locally}. LoTeNet applies MPS on small regions of images, and aggregates these representations in a hierarchical manner to retain additional spatial information. 

The use of multiple MPS operations per layer in LoTeNet results in increased computation cost compared to convolutional neural networks (CNN) with the same parameter complexity~\cite{selvan2020locally}. In this work, we attempt to reduce the computation complexity of tensor network based supervised classification models without considerable degradation in performance. To that effect, we propose to use one MPS acting on the image at multiple resolutions, instead of using several MPS per layer, resulting in the multi-layered tensor network (MLTN). Multi-layered approaches have also been  attempted in~\cite{cheng2019tree,reyes2020multi} but MLTN differs primarily in the optimisation of MPS weights and the extraction of local features, which is achieved by moving spatial information from small image neighbourhoods into the feature dimension. The resulting tensor network is  a fully linear model that performs competitively on challenging medical image classification task. 

\section{Method}

Linear decision boundaries in exponentially high dimensional spaces can be powerful; this has been the primary insight in exploring tensor networks for supervised learning~\cite{novikov2016exponential,stoudenmire2016supervised,efthymiou2019tensornetwork}. To achieve this, low dimensional data is first lifted to a high dimensional space. In this work, the linear decision boundary in high dimensional spaces is approximated using multiple layers of MPS operations. Image information in small image regions are moved to the feature dimension using the {\em squeeze} operation. These reshaped images are flattened into 1D vectors and {\em contracted} using an MPS operation to obtain intermediate representations towards yielding the final decision boundary. This output from MPS step is {\em rearranged} back into the image space forming one layer of the MLTN. A high level visualisation of the proposed model is shown in Figure~\ref{fig:lotenet}. Each of these steps are described in detail next.

\subsection{Linear models in high dimensional spaces}

Consider a 2 dimensional image (an order-2 tensor): $X \in \Rm^{H\times W}$, with $N$ pixels which is then flattened into a 1-dimensional vector $\xbf \in \Rm^N$. This flattened input image is lifted into a high dimensional space in two steps: first, a pixel-level {\em local} feature map is applied to increase the feature dimension. For any pixel, $x_j$, it is given by $\psi^{i_j}(x_j): \Rm \rightarrow \Rm^d$. Commonly used feature maps in literature include sinusoidal or intensity transformations~\cite{stoudenmire2016supervised,efthymiou2019tensornetwork,reyes2020multi}. In the second step, a joint feature map is obtained from the local feature maps by computing their tensor product resulting in an order-$N$ tensor of $d$ dimensions:
\begin{equation}
    \Phi^{i_1\dots i_N}(\xbf) = \psi^{i_1}(x_1) \otimes \psi^{i_2}(x_2) \otimes \dots \otimes \psi^{i_N}(x_N).
    \label{eq:joint}
\end{equation}
Given the high dimensional joint feature map\footnote{The tensor indices ${i_1\dots i_N}$ are dropped for ease of notation.}, $\Phi(\xbf)$, the linear  decision boundary is given by the following tensor inner product:
\begin{equation}
    f^m(\xbf) = \left ( \Theta^{m}_{i_1\dots i_N}(\xbf) \right) \cdot \left( \Phi_{i_1\dots i_N}(\xbf) \right),
    \label{eq:linModel}
\end{equation}
where $\Theta$ is an order-($N+1$) weight tensor, with output dimension, $m$, corresponding to the number of output classes\footnote{We show the indices that are being summed over as subscripts following tensor contraction notation.}. 

\begin{figure}[t]
    \centering
    \includegraphics[width=0.9\textwidth]{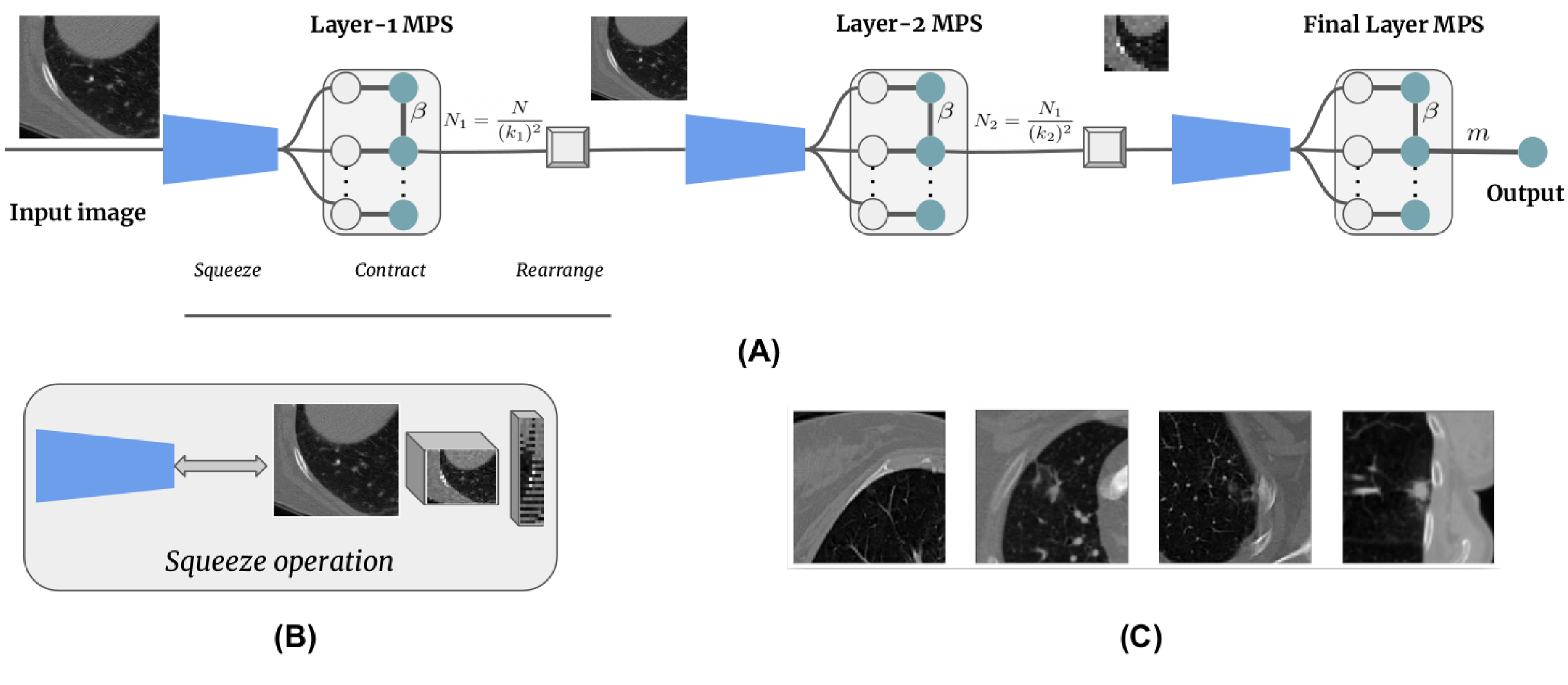}
    \caption{A: Multi-layered tensor network (MLTN) with one matrix product state (MPS) operation per layer. 
    Notice the sequence of {\em Squeeze -- Contract -- Rearrange} steps forming one layer of MLTN. 
    The output dimension of each MPS block is marked along the edges as $[N_1,N_2,m]$. B: Overview of the {\em squeeze} operation which transforms a 2D image into a vector with inflated feature dimensions. C: Four sample images from the LIDC-IDRI dataset. The first two belong to the negative class and the last two to positive class, indicating the absence and presence of tumour based on rater agreement.}
    \label{fig:lotenet}
    %\vspace{-0.25cm}
\end{figure}

The weight tensor, $\Theta$, consists of $d^N$ tunable parameters and computing the inner product in Eq.~\eqref{eq:linModel} quickly becomes infeasible with increasing $N$~\cite{efthymiou2019tensornetwork,raghav2020tensor}. One strategy to overcome this constraint is to approximate the inner product using the MPS tensor network~\cite{perez2006matrix,oseledets2011tensor,efthymiou2019tensornetwork,selvan2020locally}. MPS approximates an order-N tensor by factorising it into a chain of order-3 tensors. 
%The MPS block in Figure~\ref{fig:lotenet}-A depicts the tensor contraction based approximation to the dot product using tensor notations~\cite{penrose1971applications}. 
The weight tensor, $\Theta$, can be approximated with MPS contraction as
\begin{equation}
    \Theta^{m}_{i_1\dots i_N}(\xbf) = \sum_{\alpha_1, \alpha_2,\dots \alpha_N} A^{i_1}_{\alpha_1} A^{i_2}_{\alpha_1 \alpha_2} A^{i_3}_{\alpha_2 \alpha_3} \dots A^{m,i_j}_{\alpha_j \alpha_{j+1}} \dots A^{i_N}_{\alpha_N},
    \label{eq:mps}
\end{equation}
where  $A^{i_j}$ are the lower-order tensors. The subscript indices $\alpha_j$ are the virtual indices that are contracted. Dimension of the virtual indices is $\beta$ which is the \emph{bond dimension}. 
%The components of these intermediate lower-order tensors $A^{i_j}$ form the tunable parameters of the MPS approximation. 
The MPS approximation reduces the number of parameters from $d^N$ to \{$ d \cdot N \cdot \beta^2$\} with $\beta$ controlling the quality of these approximations. The MPS  approximation in tensor notation is indicated as the {\em contract} step in Figure~\ref{fig:lotenet}. Tensor indices are dropped in the remainder of the manuscript for ease of notation.
%For more details on MPS approximated high dimensional linear models we refer to~\cite{efthymiou2019tensornetwork,selvan2020locally}. The specific placement of the output dimension $m$ on $A^{i_j}$ in Eq.~\eqref{eq:mps} is an arbitrary choice and can be adapted during the optimisation~\citep{stoudenmire2016supervised}. 

% {\bf More details on MPS Contraction}

\subsection{Squeezed local feature maps}

The loss of spatial correlation between pixels caused by flattening 2-D images into 1-D vectors 
%in order to obtain the joint feature map in Eq.~\eqref{eq:joint} 
has been shown to hamper performance in classification tasks when dealing with images of high spatial resolution~\cite{raghav2020tensor}. In this work, we reshape small image regions so that the spatial information is moved to the feature dimension, which is similar to the {\em squeeze} operation in~\cite{dinh2016density,raghav2020tensor}. The size of the squeezed regions is controlled by the kernel stride parameter $k$ to obtain square regions of size $k\times k$. 

The transformation in dimensions due to the squeeze operation, parameterised by the stride $k$, is %given by
\begin{equation}
\psi(\cdot;k): \{ X \in \Rm^{H \times W \times d},d=1 \}  \longrightarrow \{\xbf \in \Rm^{ (N/k^2) \times d}, d= k^2\}.
\label{eq:squeeze}
\end{equation}
Note that the squeeze operation increases the feature dimension from $d=1$ to $d=k^2$. Different from the local feature maps used in literature~\cite{stoudenmire2016supervised,efthymiou2019tensornetwork}, we propose to use the squeeze operation to increase the feature dimension. 
%as a local feature map.
% However, instead of increasing the local feature map for a vector with $N$ pixels, we treat the output of squeeze operation at layer $l$ as a vector with $N\backslash (k^2)^l$ pixels and $d=k^2$ features.% and input to the MPS block. 
% Thus, the local feature maps at each layer are obtained by the squeeze operation. 

\subsection{The multi-layered tensor network}

Squeezing spatial information to the feature dimension helps in the retention of pixel correlation when compared to flattening in a single step. To further increase the retained pixel correlation, we propose to compute the decision rule in Eq.~\eqref{eq:linModel} in a hierarchical manner; this has also been attempted with different strategies in~\cite{cheng2019tree,reyes2020multi,raghav2020tensor} with varying degrees of success. In this work, we use a multi-layered approach with $L$ layers resulting in the multi-layered tensor network (MLTN):
\begin{equation}
    f_{(l)}(\xbf) = \Theta_{(l)} \cdot \Phi\left(f_{(l-1)}(\xbf)\right) \text{ for } l=1\dots L
    \label{eq:mltn}
\end{equation}
with $f_{(0)}(\xbf) = \xbf$, and $f_{(L)}(\xbf) = f(\xbf)$ which is the output tensor of order-1 and dimension $m$. The layer-wise weight tensor $\Theta_{(l)}$ in MLTN has an output dimension of $N / (k^2)^l$. As a result, the output at each layer, $f_{(l)}(\xbf)$, is a vector of dimension $N/(k^2)^l$ which is {\em rearranged} back into the image space: $({H}/{k^l}) \times ({W}/{k^l})$ before passing to the squeeze operation for layer ($l+1$). Note the exponential reduction in number of pixels between successive layers. For example, given a 128x128 input image, $k=4$ and $L=3$ the number of pixels input to the three MPS layers are: [32x32,8x8,2x2] with feature dimension $d=16$.%, as shown in Figure~\ref{fig:lotenet}-A.

Finally, note that there are no non-linear components (including in the local feature maps) in MLTN resulting in a {\em fully linear} model. MLTN with $L=3$ with the sequence of {\em Squeeze -- Contract -- Rearrange} is shown in Figure~\ref{fig:lotenet}-A and the squeeze operation itself is elaborated in Figure~\ref{fig:lotenet}-B.

{\bf Optimisation:} The parameters $[\Theta_1,\dots \Theta_L]$ in Eq.~\eqref{eq:mltn} form the weights of the model which can be learned in a supervised setting. For a given labelled training set with $T$ data points, $\mathcal{D}:\{(\xbf_1,y_1), \dots (\xbf_T,y_T)\}$, the training loss to be minimised is
\begin{equation}
    \mathcal{L}_{tr} = \frac{1}{T} \sum_{t=1}^T L(f(\xbf_i),y_i),
\end{equation}
where $L(\cdot)$ can be a suitable loss function such as cross entropy with logits for classification or mean-squared error for regression tasks. 

%The sequence of squeeze operation in Eq.~\eqref{eq:squeeze}, and MPS approximations in

% \subsection{Field of view per MPS}
% 
%\vspace{-0.2cm}
\section{Experiments}
The proposed MLTN model can be used for supervised learning tasks. We demonstrate the capabilities of MLTN for the task of image classification and compare with relevant methods to highlight its features. 

%\vspace{-0.15cm}
\subsection{Data} The LIDC-IDRI dataset consists of 1018 thoracic CT images with lesions annotated by four radiologists~\citep{armato2004lung}. We extracted $128$x$128$ px 2D slices similar to~\citep{lidc} yielding a total of $15,096$ patches. Each of these patches have annotations from four raters marking the tumour regions. These segmentation masks were converted into binary labels indicating the presence (if two or more radiologists marked a tumour) or absence of tumours (if less than two raters marked tumours), resulting in a fairly balanced dataset. All image intensities are normalised to be in $[0,1]$. 

\subsection{Experimental set-up} The proposed method was compared with three relevant methods: LoTeNet~\cite{raghav2020tensor}, single layer tensor network in~\cite{efthymiou2019tensornetwork} (denoted Tenet-X) and a multi-layered perceptron (MLP) composed of four layers. All experiments were performed using five fold cross-validation. MLTN shown in Figure~\ref{fig:lotenet}-A has two hyperparameters: the bond dimension $\beta$ and the initial kernel stride $k$ obtained from the validation performance on one of the folds, resulting in $\beta=5$ and $k=16$. The architecture for LoTeNet was the same as reported in~\cite{raghav2020tensor} with $\beta=5$. All models were implemented in PyTorch~\cite{paszke2019pytorch}, trained on a single GTX 1080 graphics processing unit with $8$GB memory using Adam optimiser~\cite{kingma2014adam} with a batch size of $512$, batch normalization~\cite{ioffe2015batch} between successive layers, for a maximum of 200 epochs and a patience of 10 epochs based on the validation accuracy. Learning rate for MLTN was $5\times 10^{-6}$ to overcome exploding gradient problem at larger learning rates, whereas for other methods it was $5\times 10^{-4}$ obtained from~\cite{selvan2020locally}. The development and training of all models in this work was estimated to produce 2.9 kg of CO2eq, equivalent to 23.9 km travelled by car as measured by Carbontracker\footnote{\url{https://github.com/lfwa/carbontracker/}}~\citep{anthony2020carbontracker}.

\begin{table}[t]
\small
\centering
    \caption{ Performance comparison on LIDC dataset. Number of parameters, computation complexity for forward propagation, computation time per training epoch (t) and area under the receiver operating characteristics (AUROC) curve (higher is better) averaged over 5-fold cross validation for all methods are reported. All three tensor network models use $\beta=5$.}
    \label{tab:oasis}
  \begin{tabular}{lccrc}
    \toprule
Models &  \# Param. ($M$) & Complexity &time (s) & AUROC\\
%& (in GB)& ($\%$)\\
    \midrule
%    {MLTN (ours)} &0.42  &$\mathcal{O}\left( \frac{\log N }{\log k^{2\cdot L}}  \cdot k^2 \cdot d \cdot \beta^2\right)$&2.8& $0.88 \pm 0.01$ \\
    {MLTN (ours)} &0.42  &$
	  \left ( \frac{\log N}{\log k^{2\cdot L}} + {L-1} \right ) \cdot k^2 \cdot d \cdot \beta^2$&2.8& $0.88 \pm 0.01$ \\
 
    %[0.8839,0.8812,0.8718,0.8765,0.8638]
    %{LoTeNet } &0.49&$\mathcal{O}\left( \frac{N \cdot L}{k^{2\cdot L}} \cdot k^2 \cdot d \cdot \beta^2\right)$ & 18.5 &  $0.87 \pm 0.01$ \\
	  {LoTeNet } &0.49&$ \left ( \frac{\log N}{\log k^{2\cdot L}} + \sum_{l=1}^{L-1}\frac{N}{k^{2 \cdot l}} \right ) \cdot k^2 \cdot d \cdot \beta^2 $ & 18.5 &  $0.87 \pm 0.01$ \\

    %[0.8992,0.8898,0.8946,0.9060,0.8869] with d=2
    %[0.8824,0.8695,0.8580,0.8811,0.8776] with d=1
    {Tenet-X} &  0.82  &$\mathcal{O}\left( N \cdot d \cdot \beta^2\right)$ &30.2& $0.82 \pm 0.01$\\
%    {Tenet-X} &  0.82  &$ N \cdot d \cdot \beta^2$ &30.2& $0.82 \pm 0.01$\\

    %[0.8537,0.8247,0.8220,0.8267,0.7940]
    {MLP (L=4)} &0.52  & $\mathcal{O}\left(N \cdot L\right)$ &3.2& $0.86 \pm 0.01$\\
    %[0.8689,0.8525,0.8772,0.8460,0.8625]
    \bottomrule
  \end{tabular}
  \vspace{-0.5cm}
\end{table}

\subsection{Results} Performance comparison across all methods is reported in Table~\ref{tab:oasis} with area under the receiver operating characteristics (AUROC) curve as the primary measure over the five folds. We noticed that MLTN and LoTeNet performed almost identically ($0.87\pm0.01$) showing a large improvement over Tenet-X ($0.82\pm 0.01$) and a smaller improvement over MLP ($0.86\pm 0.01$). The average training time per epoch for MLTN ($2.8$s) shows a clear improvement compared to LoTeNet ($18.5$s) while attaining the same performance; this difference is even clearer compared to Tenet-X ($30.2$s). Notice that MLTN, LoTeNet and MLP have comparable number of parameters ($\approx 0.5M$) compared to Tenet-X ($0.82M$).

% XXXXX THE LAST HALF HERE IS NOT "RESULTS" BUT RATHER METHODS OR DISCUSSION XXXX
% Also the number of parameters part?
% No, that belongs here.
% MOved to disc. 

%\vspace{-0.25cm}
\section{Discussions and Conclusion}
%\vspace{-0.25cm}
Local feature maps such as the sinusoidal map used in~\cite{stoudenmire2016supervised,raghav2020tensor} or the intensity based ones in~\cite{efthymiou2019tensornetwork} are used to enhance pixel level features when constructing the joint feature map. In MLTN no explicit local feature maps were used. Instead, the squeeze operation was used as an implicit feature map to retain spatial information. This has similarities with the wavelet based local feature map used in~\cite{reyes2020multi} which reduces the number of pixels by half using wavelet transforms on the image data. 
%We believe that useful local feature map representations could be learned in MLTN when trained end-to-end with the squeezed features.

The recursive formulation in Eq.~\eqref{eq:mltn} for MLTN is in contrast with LoTeNet which utilises multiple MPS blocks per layer acting on small image patches. At any layer, $l$, the field of view of an MPS block is $k\times k$ in LoTeNet, whereas for MLTN it is increased to $({H}/{k^l}) \times ({W}/{k^l})$. This could possibly improve the layer-wise representations as MPS contractions act on larger field of view. 

The computation complexity for the forward propagation of all methods are reported under the complexity column in Table~\ref{tab:oasis}, pointing to the reasons for the drastic reduction in computation time for MLTN compared to other tensor network models. The computation cost of approximating the linear decision function in Eq.~\eqref{eq:mltn} is $\left ( \frac{\log N}{\log k^{2\cdot L}} + {L-1} \right ) \cdot k^2 \cdot d \cdot \beta^2$. The cost of an MPS operation on a $k\times k$ patch is $k^2\cdot d \cdot \beta^2$. As only a single MPS is used per layer in MLTN, except the last layer, this contributes to the $(L-1)$ scaling. The final MPS layer operates on the residual spatial resolution after $(L-1)$ layers indicated as the logarithmic cost term: $ \frac{\log N}{\log k^{2\cdot L}}$. MLTN removes the linear dependency on the number of pixels ($N$) in LoTeNet (Table~\ref{tab:oasis}, row-2) and only retains the logarithmic dependence ($\log N $) due to the final MPS operation. The scaling of number of pixels by the kernel stride $k$ further reduces the computation complexity compared to Tenet-X. When compared to deep enough MLPs ($L\geq 2$), which also have an linear dependence on the number of pixels ($N$), the computation cost of MLTN due to squeeze operation ($k^2\cdot d \cdot \beta^2$) can be smaller than that of the MLP.

We observed that the modifications introduced in MLTN made it susceptible to training issues such as the exploding gradient problem. We attribute this to the large field of view of MPS operations in each of the layers. This was alleviated by using smaller learning rates ($5\times 10^{-6}$ for MLTN instead of $5\times 10^{-4}$ for LoTeNet). Reducing the depth of MLTN ($\leq 2$), data augmentation and gradient clipping~\cite{zhang2019gradient} also helped reduce the exploding gradient behaviour but these were not used in the reported experiments.

In conclusion, we proposed a modification to the locally orderless tensor network~\cite{raghav2020tensor} for supervised image classification. Instead of using multiple patch level MPS operations over successive layers, we investigate the possibility of using a single MPS operation at each layer. This modification increased the field of view of MPS operation at each layer
and reduced the number of MPS operations resulting in a substantial reduction in computation complexity (Table~\ref{tab:oasis}) without deteriorating the classification performance. The proposed formulation of MPS based tensor networks in the form of MLTN is a fully linear model and could lend itself to be applicable in more diverse settings similar to MLPs.

\subsection*{Acknowledgements}
Authors would like to thank Jon Sporring and the Medical Image Analysis group at the Machine Learning Section, for fruitful discussions.
% \newpage
\bibliographystyle{plain}
\setlength{\bibsep}{4pt}
\bibliography{selvan20}
\end{document}